\documentclass[sigconf]{acmart}

\AtBeginDocument{%
  }

\usepackage{algorithm}
\usepackage{algpseudocode}
\usepackage{xcolor}
\usepackage{multirow}
\usepackage[table,xcdraw]{xcolor}
\usepackage{colortbl} \usepackage[table,xcdraw]{xcolor}
\usepackage{graphicx}
\usepackage{subcaption}
\setcopyright{acmlicensed}
\copyrightyear{2026}
\acmYear{2026}
\acmDOI{XXXXXXX.XXXXXXX}
\acmConference[Conference acronym 'XX]{The Web Conference 2026}{June 03--05,
  2018}{Woodstock, NY}
\acmISBN{978-1-4503-XXXX-X/2018/06}

\begin{document}

\title{Forget to Generalize: Iterative Adaptation for Generalization in Federated Learning}


\settopmatter{authorsperrow=3}

\author{Abdulrahman Alotaibi}
\authornote{Both authors contributed equally to this research.}
\email{alotaibi@mit.edu}
\orcid{https://orcid.org/0000-0002-9096-2443 }
\affiliation{%
  \institution{MIT CSAIL}
  \city{Cambridge, MA}\\
 \country{USA}
}

\author{Irene Tenison}
\authornotemark[1]
\email{itenison@mit.edu}
\affiliation{%
  \institution{MIT CSAIL}
  \city{Cambridge, MA}
  \country{USA}
}

\author{Miriam Kim}
\email{miriamki@mit.edu}
\affiliation{%
  \institution{Harvard College}
  \city{Cambridge, MA}
  \country{USA}
}

\author{Isaac Lee}
\email{isaac.lee22@imperial.ac.uk}
\affiliation{%
 \institution{Imperial College London}
 \city{London}
 \country{UK}
}

\author{Lalana Kagal}
\orcid{https://orcid.org/0000-0001-8469-1993}
  \email{lkagal@csail.mit.edu}
\affiliation{%
  \institution{MIT CSAIL}
  \city{Cambridge, MA}
  \country{USA}
}


\begin{abstract}


The Web is naturally heterogeneous with user devices, geographic regions, browsing patterns, and contexts all leading to highly diverse, unique datasets. Federated Learning (FL) is an important paradigm for the Web because it enables privacy-preserving, collaborative machine learning across diverse user devices, web services and clients without needing to centralize sensitive data.  However, its performance degrades severely under non-IID client distributions that is prevalent in real-world web systems. In this work, we propose a new training paradigm — Iterative Federated Adaptation (IFA) —that enhances generalization in heterogeneous federated settings through generation-wise forget and evolve strategy. Specifically, we divide training into multiple generations and, at the end of each, select a fraction of model parameters (a) randomly or (b) from the later layers of the model and reinitialize them. This iterative forget and evolve schedule allows the model to escape local minima and preserve globally relevant representations. Extensive experiments on CIFAR-10, MIT-Indoors, and Stanford Dogs datasets show that the proposed approach improves global accuracy, especially when the data cross clients are Non-IID. This method can be implemented on top any federated algorithm to improve its generalization performance. We observe an average of 21.5\% improvement across datasets. This work advances the vision of scalable, privacy-preserving intelligence for real-world heterogeneous and distributed web systems.
\end{abstract}






\maketitle

\section{Introduction}

Federated Learning (FL) enables collaborative model training across decentralized clients without centrally pooling raw data, offering a practical foundation for building intelligent, collaborative models without compromising data privacy or sovereignty. In real-world web systems, ranging from personalized recommendation to mobile search, data distributions are inherently non-IID. When clients possess differing label mixes, feature statistics, or domain biases, naive aggregation causes client drift and poor global generalization. Classical FedAvg \cite{mcmahan2017communication} - style training often converges slowly, converges to solutions that overfit client-specific idiosyncrasies rather than learning robust abstractions, and suffer from high inter-client performance variance. Methods that improve robustness under such heterogeneity therefore represent an important step toward deployable, privacy-compliant web intelligence.

Prior FL works like FedProx \cite{fedprox} and FedNova \cite{fednova} mitigate such optimization drift by changing the optimization or aggregation rules. But they leave open the representation-level challenge: as training progresses across many rounds, later layers of the model tend to “lock in” client-specific biases and lose ability to generalize to the global distribution. Representation-level solutions such as MOON \cite{moon} or FedRep \cite{fedrep} help reduce specialization and encourage learning shared features. Although these works try to combat local drift, none explicitly re-learn parts of the model to counter cumulative drift over long federated schedules.

In this work, we propose a new paradigm for federated non-IID learning: Iterative Adaptation. Concretely, we partition the full federated training process into sequential generations. At the end of each generation, we re-initialize a fraction of the model parameters. Motivated by KE\cite{KE} and LLF\cite{zhou2022fortuitous}, we introduce two methods to select partial model parameters - (1) Random Parameter Selection: A fraction of model parameters are selected randomly across the model and (2) Later Layer Parameter Selection: The parameters of the later layers of the model (those closest to the classification head or decision boundary that learn higher-level features) are selected; to be re-initialized or reset periodically.

These two procedures enable the model to shed entrenched biases and re-learn more general features leading to improved generalization performance, especially under federated data heterogeneity. After re-initialization, the models are re-trained until the next generation. This is repeated generation-by-generation, creating a forget and evolve cycle that explicitly combats representational over-specialization and encourages continual refinement of globally relevant features (See Section \ref{why} and Algorithm \ref{alg:fl-ke-llf}). We evaluate these methods across 3 datasets - CIFAR10, MIT Indoors, and Stanford Dogs to show its effectiveness (See Section \ref{perf}). We propose iterative adaptation as a paradigm that can be implemented as a plug-and-play along with any federated aggregation algorithm to achieve generalization performance improvement under data heterogeneity beyond that brought about by the aggregation algorithm.

\section{Related Works}

\textbf{Generation-Style Learning.}
In the supervised, centralized settings, resetting specific layers has been studied in transfer learning and continual learning and shown to boost generalization performance, especially in continual learning settings. KE \cite{KE} and LLF \cite{zhou2022fortuitous} frames training as multiple generations where after training a network for one generation, some fraction of parameters are reset and retrained. SEAL \cite{seal} applies stochastic perturbations or gradient ascent in certain layers over multiple training generations, showing that perturbing layers can help avoid feature memorization and improve downstream performance. Parallel lines like the DSD (Dense-Sparse-Dense) \cite{dsd} regime, Born-Again Neural Networks \cite{bann} etc., also operate through cycles of resetting and retraining. These methods provide evidence that controlled forgetting + retraining is beneficial in centralized contexts.

\textbf{FL under Data Heterogeneity.} A substantial line of work in FL addresses non-IID data by modifying client optimization and server aggregation. Besides FedProx\cite{fedprox} and FedNova\cite{fednova}, SCAFFOLD\cite{scaffold} uses control-variates to correct biased local gradients and FedOPT\cite{fedopt} uses adaptive server optimizers to introduce momentum. Works such as FedClust \cite{fedclust} propose parameter-driven client clustering to specialize aggregation based on similarity of local models. These methods tackle heterogeneity by tailoring aggregation but do not intervene in the representational evolution inside the model. A parallel thread focuses on aligning or decoupling feature representations to cope with heterogeneous clients. Besides MOON \cite{moon} and FedRep\cite{fedrep}, methods like FedBABU\cite{babu} and FedSHIBU \cite{shibu} decouple a global extractor and local heads to limit over-specialization. Distillation-centric methods such as FedDW \cite{feddw} explicitly regularize parameter matrices via distilled relationships between class logits to mitigate heterogeneity.

\textbf{Our Approach. } By integrating insights from centralized iterative training into the federated setting, we propose a \textit{IFA} strategy which involves resetting of partial model parameters at regular training intervals followed by retraining. This iterative forget and evolve strategy is novel in the federated context and improves generalization performance.

\begin{algorithm}[t]
\caption{\textit{Iterative Adaptation for Non-IID FL}: The training process is split into $G$ generations where each generation has $C$ communication rounds with $E$ local epochs at each client. In \textit{Iterative Federated Adaptation}, at the end of each generation, $\theta_\rho$ parameters are selected and reset.}
\label{alg:fl-ke-llf}
\begin{algorithmic}[1]
\State Initialize global model $\theta^{0,0}$
\For{generation $g = 0$ to $G$}
    \For{round $c = 0$ to $C$}
        \State Sample $P$ clients from $N$ total clients
        \For{client $i \in P$ in parallel} \Comment{Client-side FL updates}
                \State Receive global model $\theta_i^{g,c} \leftarrow \theta^{g,c}$
                \State Local update of $\theta_i^{g,c}$ for $E$ epochs
                \State Return $\theta_i^{g,c}$ to the server
        \EndFor
        \State $\theta^{g,c+1} \gets \text{Aggregate}(\{\theta^{g,c_i}\}_{i \in P})$ \Comment{Any FL Aggregation}
    \EndFor
    \State Select $\theta^g_\rho$ using one of the methods below:
    \Statex \textcolor{red}{(1) Randomly select $\rho$ percent of all model parameters}
     \Statex \textcolor{blue}{(2) Select all parameters from the last $\rho$ percent of layers}
    \State Reset $\theta^{g,C}_\rho$ of $\theta^{g,C}$ to $\theta^{g,0}_\rho$
    \State $\theta^{g+1,0} \leftarrow \theta^{g,C}$
\EndFor
\State \Return 
\end{algorithmic}
\end{algorithm}

\section{Federated Iterative Adaptation}

\subsection{Method Overview}
We propose \textit{IFA}, a novel training paradigm that explicitly addresses representational drift in non-IID FL through periodic, strategic parameter reset. The fundamental insight underlying this method is that in federated environments with heterogeneous client distributions, some model parameters progressively specialize to fit client-specific data distributions, thereby accumulating \textit{representational bias}. This specialization, while potentially improving training loss on individual clients, reduces the model's ability to generalize across a broader federated population.

Formally, let us denote the training objective as minimizing the federated loss,
$$\mathcal{L}(\theta) = \sum_{i=1}^{N} w_i \mathcal{L}_i(\theta)$$
where $\mathcal{L}_i(\theta)$ is the local loss on client $i$ with weight $w_i = |D_i|/\sum_j |D_j|$. In standard federated learning, the model $\theta$ is optimized to minimize this global objective. However, under non-IID data distributions, the local gradients $\nabla \mathcal{L}_i(\theta)$ can have significant divergence, leading to what is known as \textit{client drift}. Beyond this optimization-level challenge, we identify a representational-level issue: as training progresses, some parameters learn features that are highly predictive on local data but fail to capture the invariant patterns necessary for global generalization.

Our key contribution is the introduction of an iterative \textit{forget and evolve} mechanism that periodically selects and rests a fraction of model parameters at generation boundaries. This mechanism forces the model to "un-learn" client-specific specializations while allowing other parameters to retain generalizable representations.

\subsection{Algorithmic Framework}

Algorithm~\ref{alg:fl-ke-llf} formalizes our approach. The training procedure divides the overall federated learning process into $G$ distinct generations. Within each generation $g \in \{0, 1, \ldots, G-1\}$, there are $C$ communication rounds. During each round $c \in \{0, 1, \ldots, C-1\}$, the server samples a subset $\mathcal{P}_c \subseteq \{1, 2, \ldots, N\}$ of $P$ clients from the total $N$ clients. Each client $i \in \mathcal{P}_c$ receives the current global model $\theta^{g,c}$, performs local training on their private data $D_i$ for $E$ local epochs, and returns the updated parameters $\theta_i^{g,c}$ to the server. The server then aggregates these updates using a standard FL aggregation rule (e.g., FedAvg\cite{mcmahan2017communication}): \\
$$\theta^{g,c+1} = \text{Aggregate}\left(\{\theta_i^{g,c}\}_{i \in \mathcal{P}_c}\right) = \sum_{i \in \mathcal{P}_c} \frac{|D_i|}{\sum_{j \in \mathcal{P}_c} |D_j|} \theta_i^{g,c}$$

The key innovation occurs at generation boundaries: after the final communication round $c = C$ of generation $g$ is complete, we perform \textit{targeted parameter reset}. Specifically, we select a \textit{reset fraction} $\rho \in (0, 1)$ of the model parameters to reset to that of the previous generation. Let $\theta^{g,C}_\rho$ denote the set of parameters selected for reset in generation $g$ after $C$ communication rounds. The reset operation is: $\theta^{g,C}_{\rho} \gets \theta^{g,0}_{\rho}$. The $\tilde{\rho} = 1-\rho$ fraction of parameters that were not selected, $\theta^{g,C}_{\tilde{\rho}}$ remains the same. The global model at the end of the generation will be $\theta^{g,C} \gets \{\theta^{g,0}_{\rho},\theta^{g,C}_{\tilde{\rho}}\}$

\subsection{Parameter Selection Strategies}
A critical design choice in our method is \textit{which} parameters to reset. We propose two complementary strategies that offer different trade-offs between randomness and structure. Both strategies operate within the same algorithmic framework; the choice between them can be made based on prior knowledge about architecture and task.

\subsubsection{Strategy 1: Random Parameter Selection}
In the Random Parameter Selection strategy, we uniformly and randomly select a $\rho$-fraction of all model parameters to be reset. 
In this method, each parameter in the model has an equal probability of being selected. This approach is agnostic to layer identity or parameter importance. The key advantages of this strategy are: (i) it provides unbiased selection across all model layers, preventing systematic bias toward any particular layer; (ii) it is computationally efficient, requiring only random sampling without additional analysis; and (iii) it may disrupt unexpected correlations that have emerged during training in ways that help with generalization. However, this strategy treats all parameters equally, regardless of their role in learning generalizable versus client-specific features.

\subsubsection{Strategy 2: Later Layer Selection}
The later layer Selection strategy exploits the known tendency of deep neural networks to learn increasingly specialized features in later layers \cite{bengio2013representation}. We reset all parameters only from the last $\rho$-fraction of layers (those closest to the output head). The rationale is that while earlier layers learn increasingly abstract and generalizable features across the feature hierarchy, later layers encode task and client-specific decisions \cite{zhou2022fortuitous}. By resetting only these layers, we preserve the learned abstractions in earlier layers while forcing later-layer features to be re-learned across the diverse client population. This is more targeted and reflects architectural insights.

\begin{figure}[t]
    \centering
\begin{subfigure}{\columnwidth}
        \centering    \includegraphics[width=0.3\columnwidth]{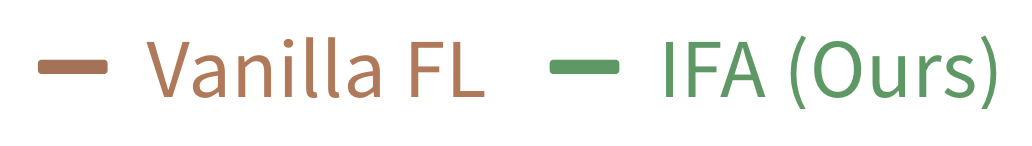}
    \end{subfigure}
    \begin{subfigure}{0.49\columnwidth}
        \centering        \includegraphics[width=\linewidth]{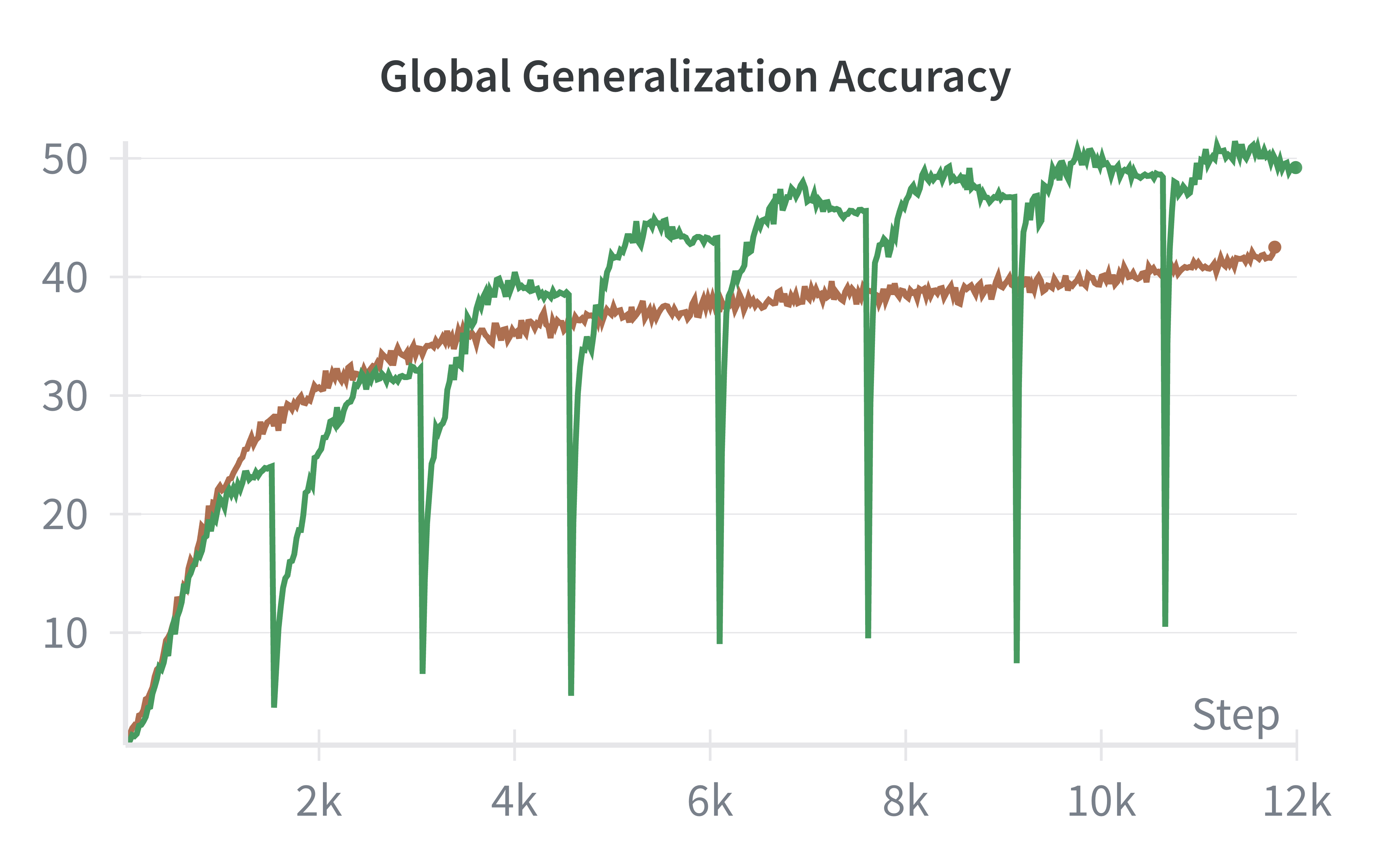}
        \label{fig:f1}
        \vspace{-0.2cm}
    \end{subfigure}
    \hfill
    \begin{subfigure}{0.49\columnwidth}
        \centering
        \includegraphics[width=\linewidth]{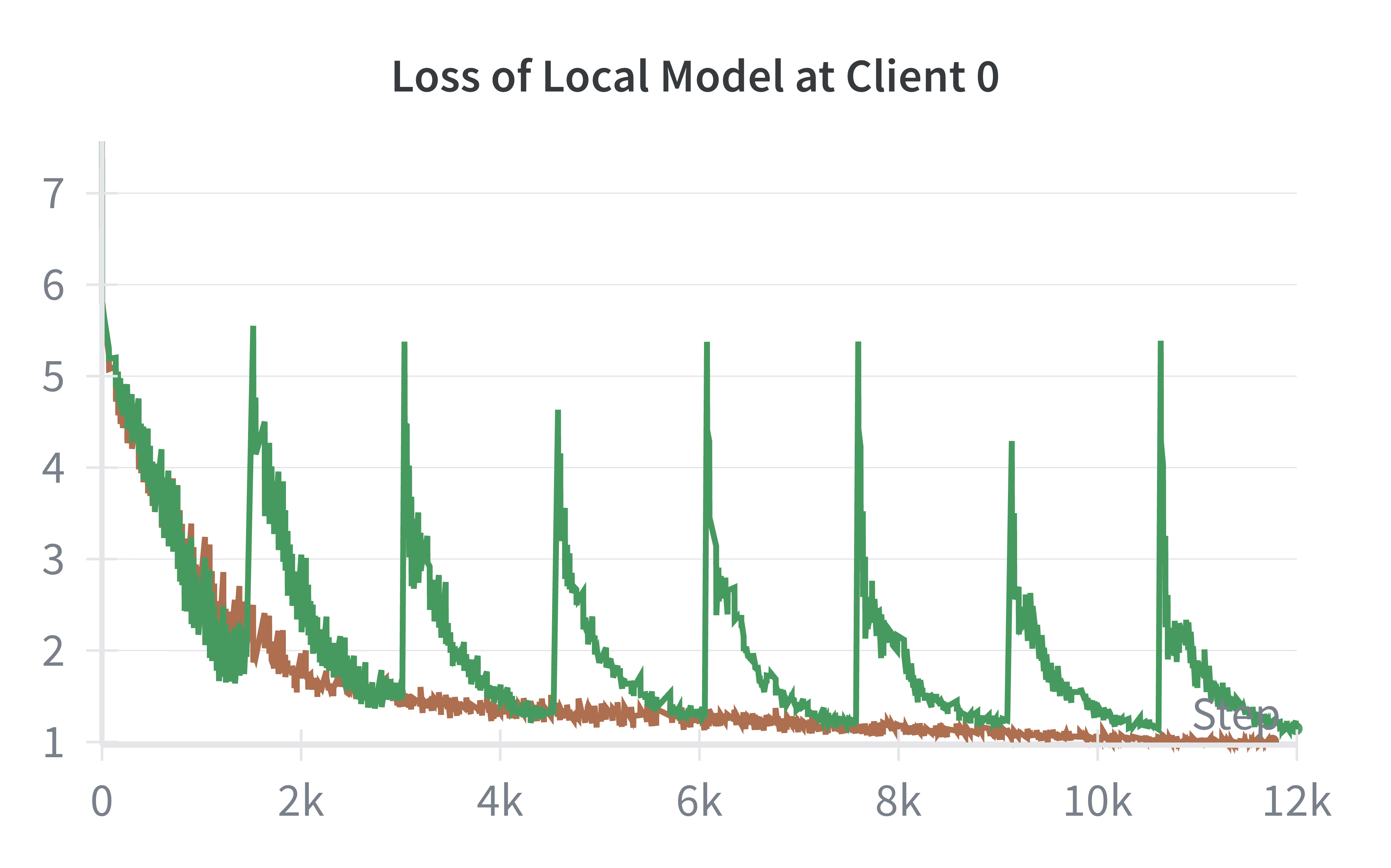}
        \label{fig:f2}
        \vspace{-0.2cm}
    \end{subfigure}
    \caption{(Left) Global model generalization performance and (Right) loss on the local model at client 0 of \textit{IFA (Ours)} and Vanilla FL (FedAVG).}
    \label{fig:onecol}
    \vspace{-0.3cm}
\end{figure}

\subsection{Why Periodic Reset Mitigates Non-IID Drift}
\label{why}
\subsubsection{Representational Drift in Federated Non-IID Settings}

Consider the optimization dynamics in federated learning with non-IID data. Each client $i$ has local data $D_i$ drawn from distribution $\mathcal{D}_i$, which may differ significantly from $\mathcal{D}_j$ for $j \neq i$. During training, each client's local optimization minimizes $\mathcal{L}_i(\theta)$, and their gradients are aggregated. Under non-IID distributions, these per-client gradients can have high \textit{variance} and \textit{divergence}. More critically for representational learning: as training progresses, the model's parameters evolve to encode compromises between diverse client objectives. In particular, parameters in later layers (closer to the prediction head) become increasingly specialized to the particular label distributions, feature statistics, and domain biases present in each client's data, also exhibiting poor transferability to other clients' distributions. This client-specificity increases with training iterations. We denote this phenomenon as \textit{representational bias}. 

We empirically observe this "client-specialization" in Figure \ref{fig:onecol}b which shows generalization accuracy and local loss on MIT CUB distributed across 10 clients. In vanilla FL, local model at client 0 (and other clients) specialize to the data at their respective clients and exhibit a lower loss than IFA indicating "client-specialization". However, over generations, IFA performs better in terms of generalization performance in comparison to vanilla FL. The "ripple-like" patterns at regular intervals represents reset of partial model parameters resulting in expected performance drop. Post-reset, the global models recover and achieve higher generalization accuracy indicating overcoming of overfitting to clients and better generalization to unseen data as further explained below.

\subsubsection{The Forget-and-Evolve Mechanism as a Regularizer}

Our periodic reset acts as an implicit regularizer that periodically \textit{resets the clock} on specialization. By resetting a fraction $\rho$ of parameters at generation boundaries, we (1) break the Specialization Momentum: Parameters that have accumulated client-specific biases over $C$ communication rounds are discarded. This prevents the model from being pulled progressively toward local client-specific optima; and (2) enable Generalizable Learning Across Diverse Clients: In generation $g+1$, the reset parameters start from the initialized state and are retrained using samples from a different subsets of clients. This reduces the inherent bias of the reinitialized parameters.

\begin{table*}[]
\caption{Accuracy of IFA with FedAVG - random selection and later layer selection, in comparison to vanilla FedAVG on CIFAR10, MIT CUB, and Stanford Dogs distributed IID and Non-IID across 10 clients. The highest accuracy for each dataset is highlighted in bold.
}
\label{acc}
\begin{tabular}{ccccccccc}
\hline
\rowcolor[HTML]{EFEFEF} 
\textbf{Method}                                                                                                                                              & \textbf{Generation}            & \multicolumn{3}{c}{\cellcolor[HTML]{EFEFEF}\textbf{Non-IID}}                   &                          & \multicolumn{3}{c}{\cellcolor[HTML]{EFEFEF}\textbf{IID}}                       \\
\textbf{}                                                                                                                                                    &                                & CIFAR10                  & Stanford Dogs            & MIT Indoors                  &                          & CIFAR10                  & Stanford Dogs            & MIT Indoors                  \\ \hline
\rowcolor[HTML]{EFEFEF} 
\cellcolor[HTML]{EFEFEF}                                                                                                                                     & Gen. 1                         &      55.16                   &                         24.62 & 25.6 &             &   76.64          &    34.74
                       & 41.94
                         \\
\multirow{-2}{*}{\cellcolor[HTML]{EFEFEF}\textbf{Vanilla FL}}                                                                                                & Gen. 10                        &     67.27                     &                       28.36   &             27.46             &                    &   77.19                             & 38.24  &        
42.84 \\ \hline
                                                                                                                                                             & 
\cellcolor[HTML]{EFEFEF}Gen. 1 & 66.93 \cellcolor[HTML]{EFEFEF} & \cellcolor[HTML]{EFEFEF} 25.9 & \cellcolor[HTML]{EFEFEF} 24.78& \cellcolor[HTML]{EFEFEF} & 77.26\cellcolor[HTML]{EFEFEF} &  \cellcolor[HTML]{EFEFEF} 34.6
 & \cellcolor[HTML]{EFEFEF} 40.45 \\

& Gen. 3                         & 65.36      & 43.11                   & 39.55                         &                                                  &                    78.5        &          41.53                &   44.18
     \\
& \cellcolor[HTML]{EFEFEF}Gen. 8 & 60.77 \cellcolor[HTML]{EFEFEF} & \cellcolor[HTML]{EFEFEF} 54.93 & \cellcolor[HTML]{EFEFEF} 44.93& \cellcolor[HTML]{EFEFEF} & 78.79\cellcolor[HTML]{EFEFEF} & \cellcolor[HTML]{EFEFEF}48.04
 & \cellcolor[HTML]{EFEFEF}43.96
 \\

\multirow{-4}{*}{\textbf{\begin{tabular}[c]{@{}c@{}} IFA (Ours):\\ Random Selection\end{tabular}}}                              & Gen. 10                        &    64.96                      &  \textbf{55.96}                        &  \textbf{48.13}                        &                        &      \textbf{79.32}                    &                     51.4     &               47.16             \\ \hline
\rowcolor[HTML]{EFEFEF} 
\cellcolor[HTML]{EFEFEF}                                                                                                                                     & Gen. 1                         &          66.37                &   26.04
                       &  23.13	                        &                          &              76.64            &  34.02
                        &               42.39           \\
\cellcolor[HTML]{EFEFEF}                                                                                                                                     & Gen. 3                         &    68.57                      &    46.22
                    &        40.82	                  &                          &        76.43                  &       55.34

                   &                      52.76    \\
\rowcolor[HTML]{EFEFEF} 
\cellcolor[HTML]{EFEFEF}                                                                                                                                     & Gen. 8                         &     69.1                     &  51.6
                         &        43.88	                  &                          & 76.43       & 60.09

                                           &  53.21                        \\
\multirow{-4}{*}{\cellcolor[HTML]{EFEFEF}\textbf{\begin{tabular}[c]{@{}c@{}}IFA (Ours):\\ Later Layer Selection\end{tabular}}} & Gen. 10                        &      \textbf{70.33}                    &        52.12                 & 44.7	                         &                          &   76.43    & \textbf{61.98}                       &                                  \textbf{55.15}                 \\ \hline
\end{tabular}
\end{table*}

This mechanism aligns with recent observations in continual learning that periodic resets, also known as \textit{catastrophic forgetting with intentionality}, can improve robustness \cite{zhou2022fortuitous}. However, our application to federated non-IID learning is novel and addresses a distinct problem setting.

\subsubsection{Agnosticity to Aggregation Methods}

Crucially, \textit{IFA} is \textit{agnostic} to the underlying federated aggregation method. Whether using FedAvg \cite{mcmahan2017communication}, FedProx \cite{fedprox}, SCAFFOLD \cite{scaffold}, or other variants, the reset mechanism operates independently at generation boundaries. This means \textit{IFA} can be composed with any existing FL algorithm to provide an additional layer of regularization against representational drift. The reset is orthogonal to the aggregation rule because it operates on parameters after they have been aggregated and before they are used in the next generation. This is an interesting direction that will be explored in the future works of IFA.

\section{Results \& Discussion}

\subsection{Evaluation Setup}
We evaluate our methods - random selection and later layer selection - against vanilla federated training (FedAVG) on image classification task using CIFAR-10, Stanford Dogs, and MIT Indoors datasets under both non-IID and IID data distributions with 10 clients. Data distributions were simulated using label skew determined by standard Dirichlet distribution where $\alpha=10000$ represented IID distribution and for Non-IID data distribution we $\alpha=0.1$ for  for CIFAR-10 and $\alpha=0.01$ for other datasets \cite{gma}. In the federated setup we simulate 10 clients with 10 generations and 66 communications rounds per generation. To simulate the equivalent for vanilla FL, we run it for 660 communication rounds with round 66 and 660 marked as generation 1 and 10 respectively. We ensure that 70\% of clients participate in each communication round with 3 local epochs per round.

\subsection{Performance Comparison}
\label{perf}
Table \ref{acc} quantitatively demonstrates the generalization improvements of IFA under both IID and non-IID regimes across three benchmarks — CIFAR-10, Stanford Dogs, and MIT Indoors—when compared to vanilla Federated Averaging (FedAVG). Compared to vanilla FL, both random parameter selection and later layer selection variants of IFA consistently achieve higher accuracy as training progresses over generations, with particularly sharp gains in the more challenging non-IID setting. For instance, after 10 generations, IFA with later layer selection surpasses FedAVG by a substantial margin on all datasets, attaining 3\% on CIFAR10 to 31\% on MIT Indoors accuracy gains. This substantiates IFA’s core premise: periodic targeted parameter resets explicitly counteract entrenched client-specific drift, yielding superior global generalization under realistic federated heterogeneity. Additionally, we note that while both IFA strategies outperform vanilla FL, the best strategy has to be chosen based on the architecture and data.



\begin{table}[]
\caption{Accuracy of IFA and vanilla FL in a federated setting on CIFAR-10 distributed Non-IID across 50 clients.}
\label{scalability}
\begin{tabular}{ccccc}
\hline
\rowcolor[HTML]{EFEFEF} 
\textbf{}                                                     & \textbf{}             & \textbf{Gen 1.} & \textbf{Gen 3.} & \textbf{Gen 10.} \\ \hline
\textbf{Vanilla FL}                                           & \textbf{}             &       65.8          &         68.08        &    68.05              \\
\rowcolor[HTML]{EFEFEF} 
\cellcolor[HTML]{EFEFEF}                                      & \textbf{Random Sel.}       &        65.62       &     65.5            &      69.61            \\
\multirow{-2}{*}{\cellcolor[HTML]{EFEFEF}\textbf{IFA (Ours)}} & \textbf{Later Layers} &       68.03          &       65.59          &       \textbf{71.81}           \\ \hline
\end{tabular}
\end{table}

\begin{table}[]
\caption{Accuracy of IFA and vanilla FL in a federated setting on CIFAR-10 distributed IID across 100 clients.}
\label{scalability}
\begin{tabular}{ccccc}
\hline
\rowcolor[HTML]{EFEFEF} 
\textbf{}                                                     & \textbf{}             & \textbf{Gen 1.} & \textbf{Gen 3.} & \textbf{Gen 10.} \\ \hline
\textbf{Vanilla FL}                                           & \textbf{}             &       72.53          &         72.53        &    72.82              \\
\rowcolor[HTML]{EFEFEF} 
\cellcolor[HTML]{EFEFEF}                                      & \textbf{Random Sel.}       &        72.38       &     74.56            &      76.03            \\
\multirow{-2}{*}{\cellcolor[HTML]{EFEFEF}\textbf{IFA (Ours)}} & \textbf{Later Layers} &       72.83          &       74.61          &       \textbf{76.67}          \\ \hline
\end{tabular}
\end{table}


\subsection{Scalability}

To assess scalability of the proposed framework, we evaluate \textit{IFA} with 50 clients. The proposed method maintains its effectiveness in these settings as shown on CIFAR-10  distributed non-IID across 50 clients in Table \ref{scalability}. Notably, the performance gains do not diminish as client population grows. Crucially, the consistent performance gains as the client population scales illustrate the robustness of IFA, indicating that targeted periodic resets remain effective for improving generalization even as federated deployment expands to larger, more realistic networks.

\section{Conclusion}
In summary, we have introduced Iterative Federated Adaptation (IFA), a principled framework for mitigating client-drift and enhancing out-of-distribution generalization in federated learning settings. Through periodic resets of randomly chosen parameters or parameters from later layers IFA enforces selective forgetting of client-specialization. IFA consistently yields substantial performance gains, an average of 21.5\% across a range of datasets and IID and Non-IID data distributions. However, while IFA effectively curbs local overfitting, an open challenge remains in adaptively determining the optimal reset schedule and layer selection in heterogeneous deployments without manual tuning. Future research focus on adaptive, data-driven strategies for selection of design choices like number of communication rounds per generation, as well as theoretical analyses that more deeply clarify the trade-offs between generalization and efficiency in large-scale, real-world federated systems. This work attempts to advance the vision of scalable, privacy-preserving intelligence for real-world heterogeneous and distributed web systems.

\bibliographystyle{ACM-Reference-Format}
\bibliography{ref_bib}

\end{document}